\documentclass[runningheads,table]{llncs}
\usepackage[T1]{fontenc}
\usepackage{graphicx}
\usepackage{multirow}
\usepackage{hyperref}
\usepackage{todonotes}
\usepackage{comment}
\usepackage{xcolor}
\usepackage{hhline}

\begin{document}
\title{The Impact of Skin Tone Label Granularity on the Performance and Fairness of AI Based Dermatology Image Classification Models 
}

\titlerunning{Impact of Skin Tone Label Granularity}
%
\author{Partha Shah\inst{1}
\and 
Durva Sankhe\inst{1}
\and 
Maariyah Rashid\inst{1}
\and 
Zakaa Khaled\inst{1}
\and
Esther Puyol-Ant\'{o}n\inst{1}
\and
Tiarna Lee\inst{1}
\and
Maram Alqarni\inst{1}
\and
Sweta Rai\inst{2}
\and
Andrew P. King\inst{1}
}
%
\authorrunning{P. Shah et al.}

\institute{School of Biomedical Engineering and Imaging Sciences, King's College London, UK \and
Dermatology Department, Kings College Hospital NHS Foundation Trust, London, UK}


%


%
\maketitle              
\begin{abstract}
Artificial intelligence (AI) models to automatically classify skin lesions from dermatology images have shown promising performance but also susceptibility to bias by skin tone. The most common way of representing skin tone information is the Fitzpatrick Skin Tone (FST) scale. The FST scale has been criticised for having greater granularity in its skin tone categories for lighter-skinned subjects. This paper conducts an investigation of the impact (on performance and bias) on AI classification models of granularity in the FST scale. By training multiple AI models to classify benign vs. malignant lesions using FST-specific data of differing granularity, we show that: (i) when training models using FST-specific data based on three groups (FST 1/2, 3/4 and 5/6), performance is generally better for models trained on FST-specific data compared to a general model trained on FST-balanced data; (ii) reducing the granularity of FST scale information (from 1/2 and 3/4 to 1/2/3/4) can have a detrimental effect on performance.
Our results highlight the importance of the granularity of FST groups when training lesion classification models. Given the question marks over possible human biases in the choice of categories in the FST scale, this paper provides evidence for a move away from the FST scale in fair AI research and a transition to an alternative scale that better represents the diversity of human skin tones.

\end{abstract}
\section{Introduction}

Artificial intelligence (AI) models based upon deep learning have shown good performance in automatically classifying skin lesions from dermatology images \cite{Estreva2017}. However, subsequent work has raised concerns about possibly biased or unfair behaviour of these models. For example, an AI model for classifying skin lesions that is trained using data from mostly lighter-skinned patients may perform better on other lighter-skinned patients compared to darker-skinned patients, i.e. its performance will be \emph{biased} in favour of lighter-skinned patients \cite{Groh2021,Daneshjou2022}. This is concerning because most available databases of dermatology images are either imbalanced by skin tone or do not report skin tone information \cite{Daneshjou2021}. Furthermore, when skin tone information is reported, it is common to do so using the Fitzpatrick Skin Tone (FST) scale \cite{Fitzpatrick1988}, which classifies skin tone into one of six categories, based on its response to ultraviolet (UV) light. The FST scale is often incorrectly conflated with race \cite{Ware2020}, is not perfectly correlated with objective assessments of skin tone \cite{Weir2024} and has been criticised for its disproportionate focus on lighter skin tones \cite{Okoji2021,Ware2020}, i.e. the categories assigned to darker-skinned patients are more “coarse” or less “granular” than those of lighter-skinned patients. Indeed, when the FST scale was first proposed, there were no categories for darker-skinned patients \cite{Fitzpatrick1988}. In other areas of healthcare AI, ``too coarse'' race information has been shown to have a significant impact on bias assessments \cite{Movva2023}.

A common approach to mitigate bias in AI models is to train or fine-tune multiple models using (entirely or mostly) protected group-specific data. If protected group information is known at inference time, these protected group-specific models can achieve better performance than a generic model trained with balanced data from all groups. Such improvements have been reported in AI models for cardiac magnetic resonance image segmentation \cite{PuyolAnton2021}, chest X-ray classification \cite{Zhang2022} and lesion classification from dermatology images \cite{Daneshjou2022}. This approach may also be preferable due to the known differences in prevalence and presentation of skin cancer between light- and dark-skinned people \cite{Bradford2009} - protected group-specific models will have the opportunity to learn group-specific features to optimise their performance. However, when training protected group-specific models it is still necessary to define the protected groups. As outlined above there are question marks over possible human biases in the most popular way of defining protected groups in dermatology, the FST scale.

Therefore, this paper goes beyond the evaluation of protected group-specific models for bias mitigation in dermatology, and evaluates the impact of the granularity of the FST scale used to define these groups.
We first investigate the training of baseline protected group-specific models based on three FST scale groups: 1/2, 3/4 and 5/6. 
We then  artificially decrease the granularity of the FST scale groups and assess the effect on model performance for different protected groups. This paper represents the first study into the effect of skin tone label granularity in AI based skin lesion classification.

The paper is organised as follows. Section \ref{sect:materials} describes the datasets utilised in the experiments. Section \ref{sect:methods} describes the training of the AI classification models. Section \ref{sect:exptresults} presents the experiments and results, which are then discussed and conclusions drawn in Section \ref{sect:discuss}.

\section{Materials and Methods}

\subsection{Datasets and Preprocessing}
\label{sect:materials}

Two publicly available datasets of clinical dermatology images were utilised: the Diverse Dermatology Images (DDI) \cite{Daneshjou2022} and Fitzpatrick 17k \cite{Groh2021} datasets.

The DDI dataset contains 656 images with FST scale information and malignant/benign diagnostic labels. The FST scale information is provided as combined categories, i.e. FST types 1 and 2 are combined into a single category, and likewise for types 3/4 and 5/6.

The Fitzpatrick 17k dataset contains 16,577 images along with associated FST scale information (types 1-6) and a 3-class diagnostic label (benign, malignant or non-neoplastic). In this paper, the target task was to classify malignant vs. benign lesions. Therefore, from the full dataset, only images with benign or malignant diagnostic labels were used and images with a non-neoplastic diagnosis were excluded. To make the FST labels consistent with those of the DDI dataset, we labelled each image with a combined FST category, i.e. 1/2, 3/4 or 5/6.

In the experiments described in Section \ref{sect:exptresults}, the data from the DDI and Fitzpatrick 17k datasets were combined. \textbf{This was crucial to achieve a sufficiently robust dataset for our analysis, particularly given the reduced sample sizes after excluding data to ensure a clear and clinically meaningful classification task and to obtain adequate representation of darker skin tones.}
For both datasets, images for which the diagnostic label did not clearly indicate whether the condition was benign or malignant were excluded. For the Fitzpatrick17k dataset, benign cases consisted of conditions such as seborrheic keratosis, dermatofibroma, warts and pilar cysts (mucous cysts were excluded), as well as a number of various nevi such as becker, congenital and halo. Malignant samples consisted of only basal cell carcinoma and squamous cell carcinoma. A similar process was followed for the DDI dataset, where benign samples included blue nevus, dysplastic nevus and melanocytic nevi. This filtering step ensured that the  diagnostic categories (i.e. benign/malignant) were used consistently across both datasets. Furthermore, samples with missing FST labels were excluded. After this filtering, we used 1,746 images from the Fitzpatrick 17k dataset (FST 1/2: 1037, FST 3/4: 574, FST 5/6: 135) and 364 images from the DDI dataset (FST 1/2: 136, FST 3/4: 147, FST 5/6: 81).

\subsection{Model Architecture and Training}
\label{sect:methods}

We employed a DenseNet-161 model architecture \cite{Huang2017} and trained all models using the Adam optimiser. The maximum number of training epochs was 100. A grid search hyperparameter optimisation was performed separately for each model to optimise batch size (16, 32, 64) and learning rate (0.0005, 0.0003, 0.0001). Model selection (for both hyperparameter choice and for choosing the best epoch) was performed using highest validation accuracy. Following \cite{Bello2024}, we applied data augmentation based on random rotations (max 90 degrees), width and height shift range of 0.15, vertical and horizontal flipping and brightness changes between 0.8 and 1.1. All images were resized to 224x224 pixels before being used for training and evaluation.
Models were trained for the binary classification task of malignant vs. benign lesion type.

\section{Experiments and Results}
\label{sect:exptresults}

\begin{figure}
    \begin{center}
    \includegraphics[width=10cm]{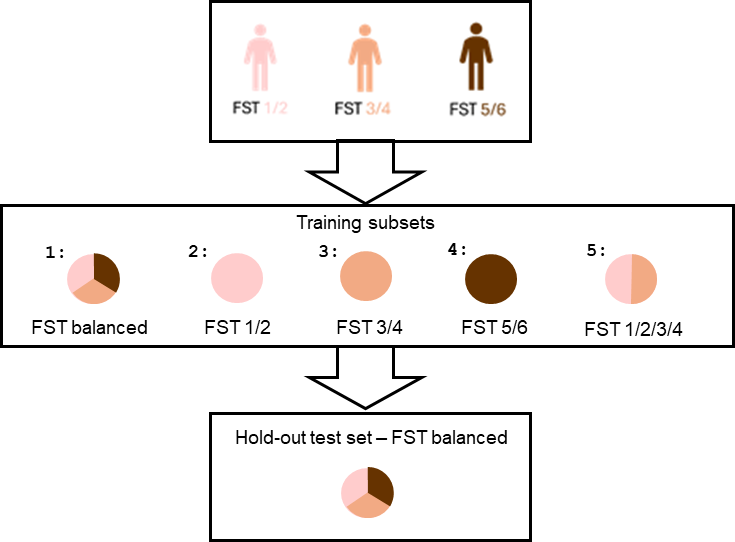}
    \end{center}
    \caption{Illustration of data subsets used for training and testing.}
    \label{fig:ddisubsets}
\end{figure}

We now introduce the experiments we performed to investigate the impact of skin tone label granularity on model performance and bias.
To carry out this investigation, we formed a number of subsets of the overall dataset as illustrated in Figure~\ref{fig:ddisubsets}. In each experiment we controlled for training set size, i.e. all compared models were trained with the same number of subjects (165 images). Each training subset was also controlled to ensure approximately equal class balance between benign and malignant lesions. The proportion of benign to malignant lesions was approximately 1:1. When training each model, the training subset was randomly split into 80\% for training and 20\% for validation. The same test set was used for all experiments, consisting of 50 FST 1/2, 50 FST 3/4 and 50 FST 5/6 images, with a 1:1 benign/malignant ratio.

Following \cite{Daneshjou2022}, we report the area under the receiver operating characteristic curve (AUC) as an evaluation metric for all models. We also report the balanced accuracy (BACC) (i.e. the arithmetic mean of sensitivity and specificity). Finally, we report the Expected Calibration Error (ECE) \cite{Guo2017}, which is a measure of uncertainty calibration, to provide a broader perspective on model performance.

All experiments were run 20 times with different random seeds and data splits and we report the mean and standard deviation of all metrics over all runs.

\subsection{Experiment 1: Baseline Protected Group-specific Models}

The first experiment aimed to determine whether training on FST-specific data, \textbf{an approach also known as stratified Empirical Risk Minimisation (ERM) \cite{Zhang2022}}, improves performance and/or fairness compared to training on mixed (FST-balanced) data (see leftmost four training subsets in Figure~\ref{fig:ddisubsets}). We evaluate all four models using AUC, BACC and ECE on each FST group individually. In addition, we quantify the fairness gap (FG), which represents the difference between the best- and worst-performing FST groups (i.e. a measurement of bias) for each metric. We compute the FG using the mean metric values over the 20 runs for the model trained with FST-balanced data and overall for the approach of training FST group-specific models.

Tables \ref{table:expt1resultsAUC}, \ref{table:expt1resultsBACC} and \ref{table:expt1resultsECE} show the results. We can see that, based on AUC, the use of FST group-specific models improves performance for all three groups (see cells highlighted in italics). The improvement for FST 1/2 was statistically significant using a two-tailed $t$-test for the AUC, BACC and ECE results. Furthermore, using FST group-specific models reduced the FG for all metrics.

\begin{figure}
    \begin{center}
    \includegraphics[width=10cm]{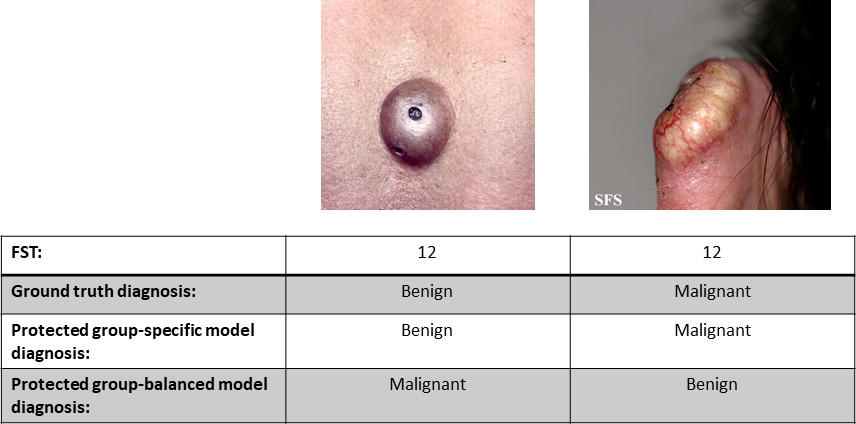}
    \end{center}
    \caption{Example test set images with FST, ground truth diagnosis and diagnoses predicted by FST-specific model and FST-balanced model.}
    \label{fig:example_images}
\end{figure}

For illustrative purposes, two sample images with their ground truth and model-predicted diagnoses are shown in Figure \ref{fig:example_images}.

From the results of this first experiment, we conclude that, even when training using data that are balanced by FST group, AI models for skin lesion classification can be biased by skin tone. This bias can be mitigated by training models using FST group-specific data, and as well as fairness improvements, this approach offers some performance improvements too.

\begin{table}[ht]
\begin{center}
\begin{tabular}{|l|cccc|}
\hline
\multirow{3}{*}{{\bf \parbox{2.2cm}{Training set}}} & \multicolumn{4}{l|}{{\bf Evaluation on test set stratified by FST}}                                                                                                    \\ \cline{2-5} 
                  & \multicolumn{4}{l|}{{\bf ~~~~~~~~~~~~~~~~~~~~~~~AUC}}                         \\ \cline{2-5} 
                  & \multicolumn{1}{c|}{{\bf FST 1/2}} & \multicolumn{1}{c|}{{\bf FST 3/4}} & \multicolumn{1}{c|}{{\bf FST 5/6}} & \multicolumn{1}{c|}{{\bf FG}} \\  \hline
                  
                  FST balanced &  \multicolumn{1}{c|}{\cellcolor[HTML]{DDDDDD} 0.83 $\pm$ 0.04} & \multicolumn{1}{c|}{\cellcolor[HTML]{DDDDDD} 0.86 $\pm$ 0.03} & \multicolumn{1}{c|}{\cellcolor[HTML]{DDDDDD} 0.93 $\pm$ 0.03} & \multicolumn{1}{c|}{0.1}  \\ \hline
                  
                  FST 1/2 & \multicolumn{1}{c|}{\cellcolor[HTML]{DDDDDD} \emph{\textbf{0.87 $\pm$ 0.03}}} & \multicolumn{1}{c|}{0.87 $\pm$ 0.04} & \multicolumn{1}{c|}{0.88 $\pm$ 0.04} & \multirow{3}{*}{0.06}  \\ \cline{2-4}\\[-3.8mm]
                  
                  FST 3/4 & \multicolumn{1}{c|}{0.84 $\pm$ 0.03} & \multicolumn{1}{l|}{\cellcolor[HTML]{DDDDDD} \emph{0.89 $\pm$ 0.03}} & \multicolumn{1}{c|}{0.88 $\pm$ 0.04} &  \\ \cline{2-4}\\[-3.8mm]
                  
                  FST 5/6 & \multicolumn{1}{c|}{0.79 $\pm$ 0.03} & \multicolumn{1}{c|}{0.88 $\pm$ 0.03} & \multicolumn{1}{l|}{\cellcolor[HTML]{DDDDDD} \emph{0.93 $\pm$ 0.02}} &   \\ \hline
\end{tabular}
\end{center}
\caption{Experiment 1 - AUC performance of classifier on different data subsets. Results shown stratified by FST group. FST = Fitzpatrick Skin Tone, AUC = area under the receiver operating characteristic curve, FG = fairness gap. Italics indicates higher performance compared to training using FST balanced data (i.e. comparison between the two shaded cells in one column). Bold = p<0.05 in two-tailed $t$-test compared to training using FST balanced data.}
\label{table:expt1resultsAUC}
\end{table}

\begin{table}[ht]
\begin{center}
\begin{tabular}{|l|cccc|}
\hline
\multirow{3}{*}{{\bf \parbox{2.2cm}{Training set}}} & \multicolumn{4}{l|}{{\bf Evaluation on test set stratified by FST}}                                                                                                    \\ \cline{2-5} 
                  & \multicolumn{4}{l|}{{\bf ~~~~~~~~~~~~~~~~~~~~~BACC}}                    \\ \cline{2-5} 
                  & \multicolumn{1}{l|}{{\bf FST 1/2}} & \multicolumn{1}{l|}{{\bf FST 3/4}} & \multicolumn{1}{l|}{{\bf FST 5/6}} & \multicolumn{1}{l|}{{\bf FG}} \\ \hline
                  
                  FST balanced &  \multicolumn{1}{c|}{\cellcolor[HTML]{DDDDDD} 0.76 $\pm$ 0.04} & \multicolumn{1}{c|}{\cellcolor[HTML]{DDDDDD} 0.81 $\pm$ 0.04} & \multicolumn{1}{c|}{\cellcolor[HTML]{DDDDDD} 0.85 $\pm$ 0.06} & \multicolumn{1}{c|}{0.09} \\ \hline
                  
                  FST 1/2 & \multicolumn{1}{c|}{\cellcolor[HTML]{DDDDDD} \emph{\textbf{0.79 $\pm$ 0.04}}} & \multicolumn{1}{c|}{0.80 $\pm$ 0.05} & \multicolumn{1}{c|}{0.82 $\pm$ 0.05} & \multirow{3}{*}{0.05} \\ \cline{2-4}\\[-3.8mm]
                  
                  FST 3/4 & \multicolumn{1}{c|}{0.78 $\pm$ 0.03} & \multicolumn{1}{c|}{\cellcolor[HTML]{DDDDDD} \emph{0.83 $\pm$ 0.04}} & \multicolumn{1}{c|}{0.81 $\pm$ 0.06} &  \\ \cline{2-4}\\[-3.8mm]
                  
                  FST 5/6 &  \multicolumn{1}{c|}{0.72 $\pm$ 0.03} & \multicolumn{1}{c|}{0.83 $\pm$ 0.04} & \multicolumn{1}{c|}{\cellcolor[HTML]{DDDDDD} 0.84 $\pm$ 0.06} &  \\ \hline
\end{tabular}
\end{center}
\caption{Experiment 1 - BACC performance of classifier on different data subsets. Results shown stratified by FST group. FST = Fitzpatrick Skin Tone, BACC = balanced accuracy, FG = fairness gap. Italics indicates higher performance compared to training using FST balanced data (i.e. comparison between the two shaded cells in one column). Bold = p<0.05 in two-tailed $t$-test compared to training using FST balanced data.}
\label{table:expt1resultsBACC}
\end{table}

\begin{table}[ht]
\begin{center}
\begin{tabular}{|l|cccc|}
\hline
\multirow{3}{*}{{\bf \parbox{2.2cm}{Training set}}} & \multicolumn{4}{l|}{{\bf Evaluation on test set stratified by FST}}                                                                                                    \\ \cline{2-5} 
                  & \multicolumn{4}{l|}{{\bf ~~~~~~~~~~~~~~~~~~~~~~ECE}}                    \\ \cline{2-5} 
                  & \multicolumn{1}{l|}{{\bf FST 1/2}} & \multicolumn{1}{l|}{{\bf FST 3/4}} & \multicolumn{1}{l|}{{\bf FST 5/6}} & \multicolumn{1}{l|}{{\bf FG}} \\ \hline
                  
                  FST balanced &  \multicolumn{1}{c|}{\cellcolor[HTML]{DDDDDD} 0.21 $\pm$ 0.04} & \multicolumn{1}{c|}{\cellcolor[HTML]{DDDDDD} 0.16 $\pm$ 0.03} & \multicolumn{1}{c|}{\cellcolor[HTML]{DDDDDD} 0.13 $\pm$ 0.05} & \multicolumn{1}{c|}{0.08} \\ \hline
                  
                  FST 1/2 & \multicolumn{1}{c|}{\cellcolor[HTML]{DDDDDD} \emph{\textbf{0.17 $\pm$ 0.03}}} & \multicolumn{1}{c|}{0.18 $\pm$ 0.05} & \multicolumn{1}{c|}{0.17 $\pm$ 0.05} & \multirow{3}{*}{0.03} \\ \cline{2-4}\\[-3.8mm]
                  
                  FST 3/4 & \multicolumn{1}{c|}{0.17 $\pm$ 0.04} & \multicolumn{1}{c|}{\cellcolor[HTML]{DDDDDD} 0.16 $\pm$ 0.03} & \multicolumn{1}{c|}{0.14 $\pm$ 0.03} &  \\ \cline{2-4}\\[-3.8mm]
                  
                  FST 5/6 &  \multicolumn{1}{c|}{0.25 $\pm$ 0.03} & \multicolumn{1}{c|}{0.17 $\pm$ 0.04} & \multicolumn{1}{c|}{\cellcolor[HTML]{DDDDDD} 0.14 $\pm$ 0.03} &  \\ \hline
\end{tabular}
\end{center}
\caption{Experiment 1 - ECE performance of classifier on different data subsets. Results shown stratified by FST group. FST = Fitzpatrick Skin Tone, BACC = balanced accuracy, FG = fairness gap. Italics indicates lower ECE (better calibration) compared to training using FST balanced data (i.e. comparison between the two shaded cells in one column). Bold = p<0.05 in two-tailed $t$-test compared to training using FST balanced data.}
\label{table:expt1resultsECE}
\end{table}

\subsection{Experiment 2: Reducing the FST Granularity of Lighter Skin Tones}

The second experiment aimed to assess the impact of reducing the granularity of the FST group labels. We randomly sampled data from the FST 1/2 and FST 3/4 groups to form a single new group (FST 1/2/3/4) with the same training/validation set sizes as the other groups. We trained using the data from this new group and compared the performance of the resulting model on the FST 1/2 and FST 3/4 test data with the performance of the original protected group-specific models (i.e. from Tables \ref{table:expt1resultsAUC}, \ref{table:expt1resultsBACC} and \ref{table:expt1resultsECE}). Therefore, in this experiment we compare the second, third and fifth models in Figure \ref{fig:ddisubsets}.

The results are shown in Tables \ref{table:expt2resultsAUCBACC} and \ref{table:expt2resultsECE}. Note that the FST 1/2 and FST 3/4 results of these tables are replicated from Tables \ref{table:expt1resultsAUC}, \ref{table:expt1resultsBACC} and \ref{table:expt1resultsECE} and are included to allow easy comparison with the FST 1/2/3/4 results.
We can see that the performance of the new `coarsened' protected group-specific model is always worse than the performance of the original protected group-specific models for all three metrics. For example, when tested on FST 1/2, the original protected group specific model (trained using FST 1/2 data) had a mean AUC of 0.87 and a mean BACC of 0.79; these figures were 0.84 and 0.75 when using the new `coarsened' model. Both of these differences were statistically significant based on a two-tailed $t$-test. The performance for the FST 3/4 group also got slightly worse for the `coarsened' model, but the differences were not significant. Likewise, the ECE values were worse for the `coarsened' model but the differences were not significant.

From this experiment, we conclude that reducing the FST label granularity when training protected group-specific models can negatively affect performance for some groups.

\begin{table}[ht]
\begin{center}
\begin{tabular}{|l|cccc|}
\hline
\multirow{3}{*}{{\bf \parbox{2.2cm}{Training set}}} & \multicolumn{4}{c|}{{\bf Evaluation on test set stratified by FST}}                                                                                    \\ \cline{2-5} 
                  & \multicolumn{2}{c|}{{\bf AUC}}                                                 & \multicolumn{2}{c|}{{\bf BACC}}     \\ \cline{2-5} 
                  & \multicolumn{1}{c|}{{\bf FST 1/2}} & \multicolumn{1}{c|}{{\bf FST 3/4}} & \multicolumn{1}{c|}{{\bf FST 1/2}} & {\bf FST 3/4}  \\ \hline
                  
                  FST 1/2/3/4 & \multicolumn{1}{c|}{\cellcolor[HTML]{DDDDDD} \emph{\textbf{0.84 $\pm$ 0.03}}} & \multicolumn{1}{c|}{\cellcolor[HTML]{DDDDDD} \emph{0.86 $\pm$ 0.02}} & \multicolumn{1}{c|}{\cellcolor[HTML]{DDDDDD} \emph{\textbf{0.75 $\pm$ 0.04}}} & \cellcolor[HTML]{DDDDDD} \emph{0.79 $\pm$ 0.03} \\ \hline
                  
                  FST 1/2 & \multicolumn{1}{c|}{\cellcolor[HTML]{DDDDDD} 0.87 $\pm$ 0.03} & \multicolumn{1}{c|}{0.87 $\pm$ 0.04}  & \multicolumn{1}{c|}{\cellcolor[HTML]{DDDDDD} 0.79 $\pm$ 0.04} & 0.80 $\pm$ 0.05  \\ \cline{2-5}\\[-3.8mm]
                  
                  FST 3/4 & \multicolumn{1}{c|}{0.84 $\pm$ 0.03} & \multicolumn{1}{c|}{\cellcolor[HTML]{DDDDDD} 0.89 $\pm$ 0.03} & \multicolumn{1}{c|}{0.78 $\pm$ 0.03} & \cellcolor[HTML]{DDDDDD} 0.83 $\pm$ 0.04 \\ \hline

\end{tabular}
\end{center}
\caption{Experiment 2 - AUC and BACC performance of classifier on coarsened subset (i.e. FST 1/2/3/4). Results shown stratified by FST group and compared to results for training using original subsets of FST 1/2 and 3/4 from Tables \ref{table:expt1resultsAUC} and \ref{table:expt1resultsBACC}.  FST = Fitzpatrick Skin Tone, AUC = area under the receiver operating characteristic curve, BACC = balanced accuracy. Italics indicates lower performance using coarsened subset compared to original subset  (i.e. comparison between the two shaded cells in one column). Bold = p<0.05 in two-tailed $t$-test compared to training using FST balanced data.}
\label{table:expt2resultsAUCBACC}
\end{table}

\begin{table}[ht]
\begin{center}
\begin{tabular}{|l|cc|}
\hline
\multirow{3}{*}{{\bf \parbox{2.2cm}{Training set}}} & \multicolumn{2}{c|}{{\bf Evaluation on test set stratified by FST}}                                                                                    \\ \cline{2-3} 
                  & \multicolumn{2}{c|}{{\bf ECE}} \\ \cline{2-3} 
                  & \multicolumn{1}{c|}{{\bf \parbox{3.2cm}{\centering FST 1/2}}} & \multicolumn{1}{c|}{{\bf FST 3/4}} \\ \hline
                  
                  FST 1/2/3/4 & \multicolumn{1}{c|}{\cellcolor[HTML]{DDDDDD} \emph{0.19 $\pm$ 0.04}} & \multicolumn{1}{c|}{\cellcolor[HTML]{DDDDDD} \emph{0.18 $\pm$ 0.03}} \\ \hline
                  
                  FST 1/2 & \multicolumn{1}{c|}{\cellcolor[HTML]{DDDDDD} 0.17 $\pm$ 0.03} & \multicolumn{1}{c|}{0.18 $\pm$ 0.05}  \\ \cline{2-3}\\[-3.8mm]
                  
                  FST 3/4 & \multicolumn{1}{c|}{0.17 $\pm$ 0.04} & \multicolumn{1}{c|}{\cellcolor[HTML]{DDDDDD} 0.16 $\pm$ 0.03} \\ \hline

\end{tabular}
\end{center}
\caption{Experiment 2 - ECE performance of classifier on coarsened subset (i.e. FST 1/2/3/4). Results shown stratified by FST group and compared to results for training using original subsets of FST 1/2 and 3/4 from Table \ref{table:expt1resultsECE} and \ref{table:expt1resultsBACC}.  FST = Fitzpatrick Skin Tone, ECE = expected calibration error. Italics indicates higher ECE (worse calibration) using coarsened subset compared to original subset  (i.e. comparison between the two shaded cells in one column). Bold = p<0.05 in two-tailed $t$-test compared to training using FST balanced data.}
\label{table:expt2resultsECE}
\end{table}

\section{Discussion and Conclusions}
\label{sect:discuss}

The FST scale is widely used as a measure of skin tone when assessing bias in AI dermatology models and when attempting to train fairer models \cite{Groh2021,Daneshjou2022,Pakzad2022}.
The main contribution of this work has been to show for the first time that the level of granularity used to record FST scale data can have a significant impact on performance of AI skin lesion classification models for different protected groups.

Specifically, Experiment 1 showed that protected group-specific models (together with inference time knowledge of protected group status) can lead to fairer outcomes and better performance for some groups.

Experiment 2 showed that reducing the granularity of the FST scale when selecting training data generally reduced performance. This is a significant finding, as the FST scale arguably already has higher granularity for lighter skin tones \cite{Okoji2021}, which represents a form of label bias.
Therefore, Experiment 2 showed that this type of label bias can negatively impact performance.

We believe that the results we have presented raise question marks over the suitability of the FST scale when training and evaluating fair AI models in dermatology.
It should also be noted that the FST scale is not actually a measure of skin tone. Rather, it categorises skin types based on their susceptibility to UV light damage \cite{Fitzpatrick1988}. It follows from this that an individual's FST scale category will never change, but their skin tone may, depending on recent exposure to UV light.
This raises further concerns over the suitability of the FST scale in fair AI work.
It may be that an alternative scale, such as the Monk skin tone scale \cite{Monk2023} or the individual typology angle \cite{Chardon1991}, which do measure skin tone and may be less inherently biased in their choice of categories, will be better suited to use in fair AI research.


Future work could extend the work presented here by evaluating the impact of FST label granularity on algorithmic bias mitigation approaches such as over-/under-sampling \cite{Kamiran2012}, loss weighting \cite{Kamiran2012}, or Group Distributionally Robust Optimisation (Group DRO) \cite{Sagawa2020}. \textbf{Additionally, exploring approaches for fair image classification that do not rely on sensitive attributes \cite{Renggli2023} or focusing on post-processing techniques like those from Ustun et al. for tabular data \cite{Ustun2019} (which could inspire image-based adaptations), might also be of interest.}

\section*{Acknowledgements}


This research was funded in whole, or in part, by the Wellcome Trust, United Kingdom WT203148/Z/16/Z.

%
%
%

\bibliographystyle{splncs04}
\bibliography{references}

\begin{thebibliography}{10}
\providecommand{\url}[1]{\texttt{#1}}
\providecommand{\urlprefix}{URL }
\providecommand{\doi}[1]{https://doi.org/#1}

\bibitem{Bello2024}
Bello, A., Ng, S.C., Leung, M.F.: Skin cancer classification using fine-tuned transfer learning of {DENSENET-121}. Appl Sci  \textbf{14} (2024)

\bibitem{Bradford2009}
Bradford, P.T.: Skin cancer in skin of color. Dermatol Nurs  \textbf{21},  170--177 (2009)

\bibitem{Chardon1991}
Chardon, A., Cretois, I., Hourseau, C.: Skin colour typology and suntanning pathways. Int J Cosmet Sci  \textbf{13}(4),  191--208 (1991)

\bibitem{Daneshjou2021}
Daneshjou, R., Smith, M.P., Sun, M.D., Rotemberg, V., Zou, J.: Lack of transparency and potential bias in artificial intelligence data sets and algorithms: A scoping review. {JAMA} Dermatol  \textbf{157}(11),  1362--1369 (2021)

\bibitem{Daneshjou2022}
Daneshjou, R., Vodrahalli, K., Novoa, R.A., Jenkins, M., Liang, W., Rotemberg, V., Ko, J., Swetter, S.M., Bailey, E.E., Gevaert, O., Mukherjee, P., Phung, M., Yekrang, K., Fong, B., Sahasrabudhe, R., Allerup, J.A.C., Okata-Karigane, U., Zou, J., Chiou, A.S.: Disparities in dermatology {AI} performance on a diverse, curated clinical image set. Sci Adv  \textbf{8}(32),  eabq6147 (2022)

\bibitem{Estreva2017}
Esteva, A., Kuprel, B., Novoa, R.A., J.~Ko, S.M.S., Blau, H.M., Thrun, S.: Dermatologist-level classification of skin cancer with deep neural networks. Nature  \textbf{542},  115--118 (2017)

\bibitem{Fitzpatrick1988}
Fitzpatrick, T.B.: The validity and practicality of sun-reactive skin types {I} through {VI}. Arch Dermatol  \textbf{124}(6),  869--871 (1988)

\bibitem{Groh2021}
Groh, M., Harris, C., Soenksen, L., Lau, F., Han, R., Kim, A., Koochek, A., Badri, O.: Evaluating deep neural networks trained on clinical images in dermatology with the {Fitzpatrick 17k} dataset. In: Proceedings of {IEEE/CVF} Conference on Computer Vision and Pattern Recognition ({CVPR}). pp. 1820--1828 (2021)

\bibitem{Guo2017}
Guo, C., Pleiss, G., Sun, Y., Weinberger, K.Q.: On calibration of modern neural networks. In: Proceedings of International Conference on Machine Learning ({ICML}) (2017)

\bibitem{Huang2017}
Huang, G., Liu, Z., van~der Maaten, L., Weinberger, K.Q.: Densely connected convolutional networks. In: Proceedings of IEEE Conference on Computer Vision and Pattern Recognition ({CVPR}) (2017)

\bibitem{Kamiran2012}
Kamiran, F., Calders, T.: Data preprocessing techniques for classification without discrimination. Knowl Inf Syst  \textbf{33},  1–33 (2012)

\bibitem{Monk2023}
Monk, E.: The {Monk} skin tone scale (2023). \doi{10.31235/osf.io/pdf4c}

\bibitem{Movva2023}
Movva, R., Shanmugam, D., Hou, K., Pathak, P., Guttag, J., Garg, N., Pierson, E.: Coarse race data conceals disparities in clinical risk score performance. In: Proceedings of Machine Learning for Healthcare ({MLHC}) (2023)

\bibitem{Okoji2021}
Okoji, U.K., Taylor, S.C., Lipoff, J.B.: Equity in skin typing: Why it is time to replace the {Fitzpatrick} scale. Br J Dermatol  \textbf{185}(1),  198--199 (2021)

\bibitem{Pakzad2022}
Pakzad, A., Abhishek, K., Hamarneh, G.: {CIRCLe}: Color invariant representation learning for unbiased classification of skin lesions. In: Proceedings of European Conference on Computer Vision ({ECCV}) (2022)

\bibitem{PuyolAnton2021}
Puyol-Ant{\'o}n, E., Ruijsink, B., Piechnik, S.K., Neubauer, S., Petersen, S.E., Razavi, R., King, A.P.: Fairness in cardiac {MR} image analysis: An investigation of bias due to data imbalance in deep learning based segmentation. In: Proceedings of Medical Image Computing and Computer Assisted Interventions ({MICCAI}). pp. 413--423 (2021)

\bibitem{Renggli2023}
Renggli, C., Smith, D.B., Gola, H.M., Kindermans, P.J.: Do we need training data? towards fairness in computer vision with no-reference metrics. arXiv preprint arXiv:2309.05148  (2023), \url{https://arxiv.org/abs/2309.05148}

\bibitem{Sagawa2020}
Sagawa, S., Koh, P.W., Hashimoto, T.B., Liang, P.: Distributionally robust neural networks for group shifts: On the importance of regularization for worst-case generalization. In: Proceedings of International Conference on Machine Learning ({ICML}) (2020)

\bibitem{Ustun2019}
Ustun, F.O., Laumann, J.R., Smith, A.D.: Fairness without demographics in repeated decisions. In: Proceedings of the 36th International Conference on Machine Learning (2019), \url{https://proceedings.mlr.press/v97/ustun19a.html}

\bibitem{Ware2020}
Ware, O.R., Dawson, J.E., Shinohara, M.M., Taylor, S.C.: Racial limitations of {F}itzpatrick skin type. Cutis  \textbf{105}(2),  77--80 (2020)

\bibitem{Weir2024}
Weir, V.R., Dempsey, K., Gichoya, J.W., Rotemberg, V., Wong, A.K.I.: A survey of skin tone assessment in prospective research. npj Digit Med  \textbf{7}(191) (2024)

\bibitem{Zhang2022}
Zhang, H., Dullerud, N., Roth, K., Oakden-Rayner, L., Pfohl, S., Ghassemi, M.: Improving the fairness of chest {X}-ray classifiers. In: Proceedings of Conference on Health, Inference, and Learning. pp. 204--233 (2022)

\end{thebibliography}

\end{document}